\documentclass[twocolumn]{article}

\usepackage{comment}

\usepackage{amsmath}
\usepackage{amssymb}
\usepackage{graphicx}
\usepackage{hyperref}
\usepackage{multirow}

\DeclareMathOperator*{\argmax}{arg\,max}

\newcommand{\OWQEext}{Online Weighted Q-Ensemble}

\title{Online Weighted Q-Ensembles for Reduced Hyperparameter Tuning in Reinforcement Learning}
\author{Renata Garcia and Wouter Caarls\\Pontifical Catholic University of Rio de
Janeiro\\renata.garcia.eng@gmail.com, wouter@ele.puc-rio.br}
\date{}

\begin{document}

\maketitle

\abstract{Reinforcement learning is a promising paradigm for learning robot control, allowing complex control policies to be learned without requiring a dynamics model. However, even state of the art algorithms can be difficult to tune for optimum performance. We propose employing an ensemble of multiple reinforcement learning agents, each with a different set of hyperparameters, along with a mechanism for choosing the best performing set(s) on-line. In the literature, the ensemble technique is used to improve performance in general, but the current work specifically addresses decreasing the hyperparameter tuning effort. Furthermore, our approach targets on-line learning on a single robotic system, and does not require running multiple simulators in parallel.

Although the idea is generic, the Deep Deterministic Policy Gradient was the model chosen, being a representative deep learning actor-critic method with good performance in continuous action settings but known high variance. We compare our online weighted q-ensemble approach to q-average ensemble strategies addressed in literature using alternate policy training, as well as online training, demonstrating the advantage of the new approach in eliminating hyperparameter tuning. The applicability to real-world systems was validated in common robotic benchmark environments: the bipedal robot half cheetah and the swimmer. Online Weighted Q-Ensemble presented overall lower variance and superior results when compared with q-average ensembles using randomized parameterizations.}

\maketitle

\section{Introduction}
    \label{sec:introduction}
    \label{sec_introduction}
    
    Reinforcement learning (RL) is based on a mathematical framework known as a Markov Decision Process (MDP)~\cite{sutton_intro}.
	Control policies for many common domains such as motor control~\cite{liu2021deep}, treatment planning~\cite{watts2020optimizing} and disease spread prediction~\cite{khalilpourazari2021designing} can be optimized under this framework, by maximizing a reward signal in order to achieve a goal.
	RL can be distinguished from other optimal control approaches such as dynamic programming by the fact that an a priori model of the environment is not required.
	Recent advances in reinforcement learning using deep neural networks \cite{kai2017deep_reinforcement_learning_survey} allow it to be applied to increasingly complex environments.
	In robotics, this means optimizing more complex motions, requiring the simultaneous actuation of many joints. In these environments, the state space (joint positions and velocities) is continuous.
   
    For discrete actions, the DQN (Deep Q-Learning) algorithm \cite{mnih2015dqn} has shown good performance results for Atari games, while for continuous actions such as robotics, the DDPG (Deep Deterministic Policy Gradient)~\cite{lillicrapcontinuous} method and its variants TD3~\cite{fujimoto2018td3errorinactorcriticmethods} and SAC~\cite{haarnoja2018soft_sac} are more suitable.
    All these solutions utilizing deep learning algorithms need fine-tuning of their hyperparameters to converge.
    One approach is to perform grid search~\cite{Oliveira2021cba} or genetic search~\cite{fernandez2018} to automate the tuning.
    These algorithms have a high computational cost, and are difficult to apply if the optimization is to take place in the real world instead of in simulation.
    
    Alternatively, an \emph{ensemble} of different sets of hyperparameters can be trained to decrease the hyperparameter tuning effort~\cite{Oliveira2021icaart}.
    Ensembles were first used in RL before the advent of deep learning techniques to increase performance~\cite{wiering2008ensemble,hans2010ensembles,duell2013ensembles}, and recent efforts have shown that ensemble aggregations of deep neural networks perform better than a single algorithm as well~\cite{wu2020ddpgensemble}.
    Some proposed ensemble aggregations have additional parameters, which add more variables to be fine-tuned, while others present a population-based approach to improve performance~\cite{jung2020population}.
    
    While ensembles in deep RL thus demonstrated good results when used to increase performance, there have been few studies of the behavior of RL ensembles with different hyperparameters, aimed at reducing the tuning effort.
    The history-based framework in~\cite{Oliveira2021icaart} is the first study to seek optimized techniques of ensemble deep reinforcement learning to decrease the hyperparameter tuning effort, where differently parameterized DDPG policies are trained online in a MuJoCo environment \cite{todorov2012mujoco}.
    However, it was based purely on action aggregation, without taking into account the value function those actions are based on.
    
    This article aims to improve the ensemble aggregation strategy by using the weighted mean of value functions trained using different hyperparameters.
    This mean is used to choose an action from among corresponding control policies trained using those hyperparameters.
    In previous work, such a maximum Q-average ensemble \cite{hans2010ensembles,anschel2017averaged} has been shown to work, although with unweighted averaging.
    We introduce a weighted average in order to deal with the larger expected variance between the outputs of vastly differently parameterized value functions.
    
    The Deep Deterministic Policy Gradient (DDPG) algorithm was chosen to validate the model, as it represents a family of state-of-the-art algorithms known for their good learning performance in continuous action settings \cite{shen20b_deep_reinforcement_learning_with_robust_and_smooth_policy}.
    However, the approach should be applicable to any off-policy actor-critic method.
    We believe it is the first use of value function ensembles specifically targeted at eliminating hyperparameter tuning.
    Additionally, it does not require parallel environments \cite{Motehayeri_addpg_parallel_envs}, and as such is applicable to real-world systems.

    This article is organized in~\ref{sec_conclusion} sections.
	Section~\ref{sec_background} presents the DDPG algorithm and Q-Average ensemble method, while
	Section~\ref{sec_wce} presents the novel Online Weighted Q-Ensemble. The experiments are defined in Section~\ref{sec_experiment} and their results discussed in Section~\ref{sec_results}.
	Finally, 
         Section~\ref{sec_conclusion} presents the conclusion.
    
\section{Background}
    \label{sec_background}
    
    RL is a type of machine learning technique that enables an agent to learn in an interactive environment by trial and error using feedback from its own actions and experience, seeking to maximize rewards.
    Every time step $t$, an action $a \in \mathcal{A}$ at state $s \in \mathcal{S}$ is taken in the environment, which returns a reward $r \in \mathbb{R}$ and results in the next state $s' \in \mathcal{S}$. The goal is to find the control policy $\pi(a\mid s)$ that gives the probability of taking action $a$ in state $s$ that maximizes the expected sum of future rewards, also called the \emph{return} $R$:
    \begin{equation}
        R_t = \sum_{i=0}^\infty \gamma^i r_{t+i+1},
    \end{equation}
    where $\gamma$ is a \emph{discount factor} introduced to avoid infinite returns.
    
	In order to scale RL techniques, Deep RL learns its own state representation and can thereby solve complex problems allowing the application in many domains of decision making tasks such as healthcare, robotics, smart grids and finance~\cite{franccois2018introduction}.
            
	\subsection{Deep Deterministic Policy Gradient (DDPG)}
	\label{sec_deep_deterministic_policy_gradient}
	
        \begin{figure}
            \centering
    	    \includegraphics[width=0.4\textwidth]{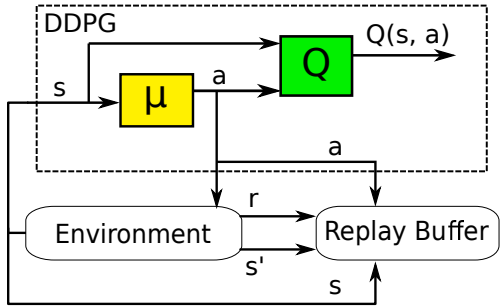}
            \caption{Deep Deterministic Policy Gradient (DDPG) algorithm \cite{lillicrapcontinuous}, its interaction with the environment and the storage of transactions in the Replay Buffer.}
            \label{fig_ddpg_model}
        \end{figure}
        
        DDPG~\cite{lillicrapcontinuous} is based on the Deterministic Policy Gradient (DPG), which was one of the first model-free and off-policy actor-critic algorithms for continuous state and action spaces~\cite{silver2014dpgdeterministic}. DDPG, illustrated in Figure~\ref{fig_ddpg_model}, is an extension of DPG that uses deep neural networks to approximate the actor and critic.
        The actor approximates a deterministic policy $\mu(s;\theta)$ with weights $\theta$, such that $\pi(a\mid s) = 1 \text{ iff } a = \mu(s; \theta)$. The critic estimates the expected return of $\mu$ by approximating the action-value function
        \begin{equation}
            Q^\pi(s,a) = \mathbb{E}_\pi \left[ R_t \mid  s_t = s, a_t = a \right]
        \end{equation}
        using a neural network, as in deep Q-learning \cite{mnih2014recurrent}.
         
        The critic network, with weights $\zeta$, is updated to minimize the loss function
     	\begin{equation}
          	\begin{aligned}
            \label{eq_q_min_squared_td_error}
        	L(\zeta) &= \mathbb{E}_{(s,a,r,s')}\left[\left(Q(s, a;\zeta) - y_t \right)^2\right]\\
        	y_t &= r_t + \gamma Q(s', \mu(s';\theta') ; \zeta'),
        \end{aligned}
     	\end{equation}
        where $\theta'$ and $\zeta'$ are Polyak-averaged versions of the main network parameters, also called \emph{target networks}, used to stabilize the learning. They are updated at given intervals using an averaging parameter $\tau$:
        \begin{align}
            \label{eq_soft_update_target_network_Q}
            \zeta' &= (1-\tau) \cdot \zeta' + \tau \cdot \zeta\\
            \label{eq_soft_update_target_network_M}
            \theta' &= (1-\tau) \cdot \theta' + \tau \cdot \theta .
        \end{align}        
        
        The actor update takes a step in the positive gradient criteria of the critic with respect to the actor parameters, given by the chain rule
        \begin{equation}
            \label{eq_actor_update}
            \begin{aligned}
                \nabla_{\theta} J &= \mathbb{E}_{s_t \sim \rho^{\beta}} \left [ \nabla_{\theta} Q (s,\mu(s;\theta);\zeta) \rvert _{s=s_t} \right ],\\
                &= \mathbb{E}_{s_t \sim \rho^{\beta}} \left [ \nabla_{a} Q (s,a;\zeta) \rvert_{s=s_t, a=\mu(s_t)} \nabla_{\theta} \mu(s;\theta)\rvert_{s=s_t})   \right ],
            \end{aligned}
        \end{equation}
        thus moving the control policy in the direction of increased returns. In Eq.~(\ref{eq_actor_update}), $J$ represents the expected return over the start distribution and $\rho^\beta$ is an exploratory stochastic behavior policy.
        
        An experience replay buffer $\mathcal{R}$ stores observed transitions $(s,a,r,s')$, in order to learn from past experience. Updates are performed using a random minibatch
     	$\mathcal{D}$ sampled from $\mathcal{R}$, used to temporally decorrelate the observations.
        
        The behavior policy $\rho^{\beta}$ in Eq.~(\ref{eq_actor_update}) is derived from the actor by adding noise, $\rho^\beta \sim\mu(s;\theta) + \mathcal{N}$, 
        to improve the exploration.
        The $\mathcal{N}$ uses the Ornstein-Uhlenbeck process~\cite{uhlenbeck1930theory} for physical environments that have momentum to generate time-correlated exploration for increased efficiency ($\mathcal{N}$$\theta$ = 0.15 and $\mathcal{N}$$\sigma$ = 1).
        The Ornstein-Uhlenbeck process models the velocity of a Brownian particle with friction, which results in temporally correlated values centered around zero~\cite{lillicrapcontinuous}.
        
    \subsection{Q-Average Aggregation}
    \label{sec_average_ddpg}
    
        In Ensemble RL, multiple value functions and/or policies are learned at the same time, and their actions aggregated to determine the ensemble action. One of the best ensemble aggregation methods developed in previous work is value function averaging.
        This has been successful in both regular reinforcement learning in discrete environments~\cite{sun1999multi,ernst2005tree} as well as deep RL in continuous action spaces~\cite{huang2017learning}.
        
        The Actor-Critic Ensemble (ACE~\cite{huang2017learning}) introduced the use of the Deep Deterministic Policy Gradient (DDPG) algorithm for ensembles. 
        At inference time, the best action is selected from all actors running in parallel, each using the outputs of all critic networks which are combined by taking the average.
        This work showed significant improvement in the performance of DDPG in a bipedal walking environment; it increased the learning speed and lowered the number of falls.

\section{Online Weighted Q-Ensemble}   \label{sec_wce}

    Our Q-ensemble model builds upon the Actor-Critic Ensemble method, by weighing the critics' predictions.
    Such a weighing, similar to conventional classifier Boosting~\cite{freund1996experiments}, aims to emphasize the input of the critics that better estimate the return when selecting the ensemble action.
    Considering that in our case the DDPG ensemble hyperparameters are chosen randomly with only little user input, it is important that critics with bad performance do not destabilize the final policy. Our critic weight update is therefore designed to decrease the weight of such critics.
    
    \subsection{Inference}

\begin{figure*}
        \centering
        \includegraphics[width=\textwidth]{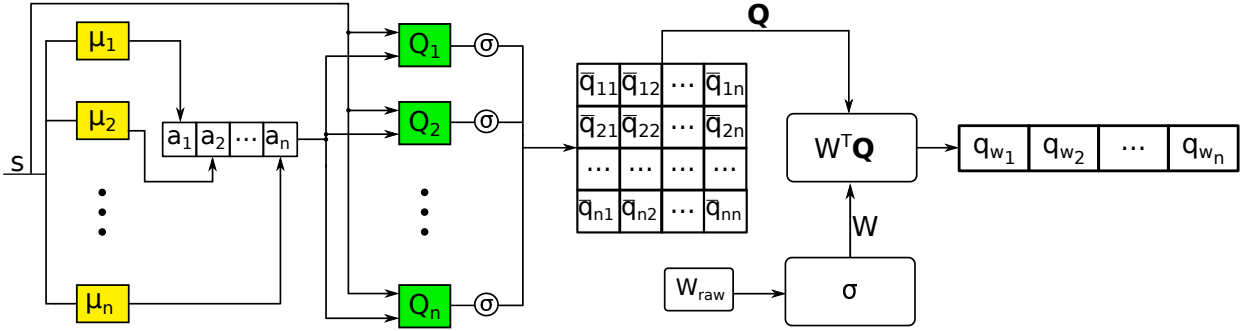} 
        \caption{Online Weighted Q-Ensemble. Each critic $Q_i$ evaluates the actions suggested by all actors $\mu_j$, and the resulting values are normalized using the softmax function $\sigma$. The final Q-value $q_{\mathrm{w}_j}$ of each action $a_j$ is the sum of the critics' values $\{\bar{q}_{1j}, \bar{q}_{2j}, \cdots, \bar{q}_{nj}\}$, weighted by their respective weights $w_i$.}
        \label{fig_wce_layer}
    \end{figure*}

    Figure \ref{fig_wce_layer} presents the process by which Q values are calculated according to the Online Weighted Q-Ensemble model. The ensemble is composed of $n$ DDPG agents, each composed of a critic $Q_i= Q(\cdot,\cdot;\zeta_i)$ and an actor $\mu_j=\mu(\cdot; \theta_j)$, $i, j \in 1 \ldots n$.
    At state $s$ at a given time step $t$, each actor $\mu_j$ calculates an action $a_j = \mu(s_t;\theta_j)$.
    Subsequently, each of the actions is evaluated by all critics, generating the corresponding Q-values $Q(s_t, a_j; \zeta_i)$. The generation of these values results in a matrix $\mathbf{Q}$ with elements $q_{ij}$.

    The DDPG networks are updated independently, since they use different hyperparameters.
    This creates a challenge when analyzing the $\mathbf{Q}$ matrix to select the ensemble action because their Q-value magnitudes may not be directly comparable.
    We propose the use of a softmax function $\sigma$ to normalize the values of a critic for the different actions:
    
    \begin{equation}
        \sigma(q_{ij}) = \frac{e^{q_{ij}}}{\sum^n_{k=1} e^{ q_{ik}}}.        \label{eq_wce_softmax_function}
    \end{equation}

    The $\sigma(q_{ij})$ normalizes the q-values $q_{ij}$ over all q-values related to value function $Q_i$, resulting in $\bar{q}_{ij}$. This procedure is equivalent to interpreting each critic as defining a softmax policy over the suggested actions. Averaging these action probabilities does not suffer from the incomparability of Q-values.

    In the final step, the Q values are combined across critics using weighted averaging.
    The raw critic weights $W_\mathrm{raw}$ are normalized with the same softmax function $\sigma$ to ensure they form a proper distribution, resulting in $W$.
    Then, we use the weighted average to calculate the critic ensemble prediction for all actions
    
    \begin{equation}
        q_\mathrm{w} = W^T\mathbf{Q},
        \label{eq_weighted_average}
    \end{equation}
    
    where $W = \frac{1}{n}\mathbf{1}$ results in standard Q value averaging. Due to the normalization of the $\mathbf{Q}$ matrix, this procedure is equivalent to weighted Boltzmann addition~\cite{wiering2008ensemble}.
    The final ensemble action, $a_t$, is the one with the highest probability:
    
    \begin{equation}
        a_t = \mu_z(s_t),
        \label{eq_wce_argmax_action}
    \end{equation}
    where $z=\argmax q_\mathrm{w}$.

    \subsection{Online Training of Weights}
    \label{sec_wce_training}
    
        All DDPG agents are trained in a single environment, using a shared replay buffer.
        The behavior policy $\upsilon^\beta$ is either derived from the ensemble action (online training) or from each actor in sequence on a per-episode basis (alternate training).
        The former can be expected to learn faster, while the latter ensures at least some near on-policy transitions for all actors, which may increase robustness.
        
        The raw weights, $W_\mathrm{raw}$, are initialized uniformly, and passed through a softmax layer before being used for the weighting. Therefore, at the beginning of the training, the critic weights ${W}$ remain close to the uniform distribution. During training, we minimize the temporal difference (TD) error of the critic ensemble by minimizing the loss
        
     	\begin{equation}
          	\begin{aligned}
            	\mathcal{L}({W_\mathrm{raw}}) &= \sum_{(s, a, r, s') \in \mathcal{D}} \sum_{i=1}^{n} w_i \delta_i^2\\
            	\delta_i &= \quad r + \gamma Q(s',\mu(s';\theta'_i);\zeta'_i) - Q(s,a;\zeta_i).
          	\end{aligned}
            \label{eq_wce_minimize_tderror}
        \end{equation}
        
        over the weight parameters. Minimizing Eq. (\ref{eq_wce_minimize_tderror}) reduces the weights of the critics with higher squared TD error $\delta$, which can be assumed to have a worse value function prediction, see Eq.~(\ref{eq_q_min_squared_td_error}).
        Since $\sum_i w_i = 1$, due to the softmax function applied to $W_\mathrm{raw}$, this necessarily increases the better critics' weights.
        
        Note that although the Q values are normalized for action selection during inference, the TD errors calculated in Eq.(\ref{eq_wce_minimize_tderror}) during training are not.
        As such, we do not optimize hyperparameters that inherently greatly influence the Q values, specifically the discount rate $\gamma$ and reward scale.
        
        The code used in the {\OWQEext} can be found online\footnote{\url{https://github.com/renata-garcia/wce_ddpg}}.

    \subsection{Performance Measure}
    \label{owqe_performance_measure}
    
        We introduce a performance metric to compare the different forms of aggregation between different environments. It has the property of being invariant to both constant addition and multiplication of the reward function, which allows some measure of robustness in comparing environments that have performance values at different scales.

        As such, to measure the overall performance of the aggregations used, the \emph{average relative regret} is calculated as
        
        \begin{equation}
            \label{eq_wce_average_relative_regret}
            \mathcal{R}(k) = \sum_e \sum_g \left\lvert \frac{\max_l p_{e,g,l} - p_{e,g,k}}{\max_l p_{e,g,l} - \min_l p_{e,g,l}} \right\rvert,
        \end{equation}
        \\
        where $p_{e,g,k}$ is the performance of aggregation strategy $k$ in environment $e$ for ensemble group $g$ (an ensemble group is a specific way of constructing the ensemble). This metric measures how much worse a certain aggregation strategy is, relative to the best aggregation for that experiment. Higher regrets mean worse overall performance.

\section{Experiments} \label{sec_experiment}

    To validate the model, we test its performance by ablating the two differences with respect to Q-value averaging: using a weighted average, and using Bolzmann addition. This results in the following combinations:
    
    \begin{itemize}
        \item \textit{Softmax TDError}: uses the model presented in Section~\ref{sec_wce};
        \item \textit{TDError}: skips the softmax normalization of the Q-values presented in Eq.~(\ref{eq_wce_softmax_function});
        \item \textit{Softmax Average}:  maintains $W=\frac{1}{n}\mathbf{1}$, but otherwise implements the model of Section~\ref{sec_wce};
        \item \textit{Average}: standard Q-value averaging~\cite{huang2017learning}.
    \end{itemize}
    
    The average policy ensemble, with DDPG agents independently trained and no q-ensemble, recently presented good performance with 3 fine-tuned hyperparameter sets~\cite{wu2020ddpgensemble} in a single environment 2D robot arm simulator.
    Based on this result, one ensemble group with $3$ fine-tuned DDPG (\textit{3 Good}) instances is used to validate the model.
    
    However, the Online Weighted Q-Ensemble seeks to minimize the effort of fine-tuning in an ensemble, and to that end $3$ more ensemble groups were created, mixing good (fine-tuned) and bad (not fine-tuned and non-converging) DDPG hyperparameters: \textit{1 Good and 1 Bad}; \textit{1 Good and 3 Bad}; and \textit{1 Good and 7 Bad}.
    
    In addition, two types of training mode are used in order to expand the validation of model.
    In the alternate training mode, at the beginning of each training episode, the policy is chosen alternately between each of the algorithms of the ensemble, as in~\cite{wu2020ddpgensemble}.
    In the online training mode and in the testing phase of the ensemble, the ensemble action is chosen at each step of the episode.
    
    As environments, we chose two simple control problems, and two harder robotics tasks to evaluate scalability. In all cases, the episodes start at the resting point of the environment, the observations of the environments are in trigonometric format and there are $1000$ observation steps before starting training. The ranges of the hyperparameters are given in Table~\ref{tbl_hyperparameters_explained}.
    	\begin{table*}
    		\centering
    		\caption{Table with hyperparameter descriptions used to build each parameterization of the ensemble.}
    		\footnotesize
    		\label{tbl_hyperparameters_explained}
    		\begin{tabular}{l|l|l}
    			\hline
    			\textbf{Hyperparameters} & \textbf{Value} & \textbf{Description} \\ \hline \hline
    			discount factor & 0.99 & Discount factor used in the Q-learning \\
    			& & update. \\ \hline
    			reward scale & 0.01 & Scaling factor applied to the environ-\\
    			& & ment's rewards. \\ \hline
    			soft target & 0.01 & Update rate of the target network\\
    			update rate & & weights. \\ \hline
    			update interval & 10 or 100 & Number of steps, or frequency \\
    			& &  with which the soft target update \\
    			& & is applied. \\ \hline
    			learning rate & 0.001 or 0.0001 & Update rate used by AdamOptimizer. \\ \hline
    			replay steps & 64, 128 or 256 & Total number of training samples \\
                & & per timestep. \\ \hline
                minibatch size & 16, 64 or 128 & Number of training samples per minibatch. \\ \hline
    			layer size & 50, 100, 200, 300 & Number of neurons in regular densely-\\
    			&  or 400 & connected NN layers \\ \hline
    			activation function & relu or softmax & Activation function of the hidden layers. \\ \hline
    			replay memory & 1000000 & Size of the replay memory array that \\
    			size & & stores the agent's experiences in the \\
    			& & environment. \\ \hline
    			observation steps & 1000 & Observation period to start replay \\
    			& & memory using random policy. \\ \hline
    		\end{tabular}
    	\end{table*}
    

    The specific environments used are the Inverted Pendulum Swing-up (2 state variables) and Cart-Pole environments (4 state variables) from the Generic Reinforcement Learning Library (GRL) \footnote{GRL Library (Generic Reinforcement Learning Library) (\url{https://github.com/wcaarls/grl}).}, and half cheetah v2 (17 state variables) and swimmer v2 (8 state variables) from the OpenAI Gym framework \cite{Brockman2016} with the MuJoCo environments \cite{todorov2012mujoco}. 
    
    Swimmer v2 was used as a final validation for hyperparameter randomization. For this environment, 30 random configurations of ensembles formed with 8 parameterizations were trained once. The network architecture and the hyperparameters were randomly generated within the limits given before (Table~\ref{tbl_hyperparameters_explained}).
    
    In order to measure the overall performance of the aggregations used, the average relative regret in Eq.~(\ref{eq_wce_average_relative_regret}) is calculated over the first three environments and all four ensemble groups. 
    	
    \section{Results}
    \label{sec_results}
    
    In the this section, the average performances presented are calculated based on the cumulative rewards of the last 20 episodes in each run, and the $95\%$ confidence interval is calculated over 30 runs of each configuration for the simple control problems and 10 runs for the half cheetah v2.
    
    \subsection{Performance}
    
\begin{figure*}[tb!]
            \centering
            \includegraphics[width=.42\textwidth]{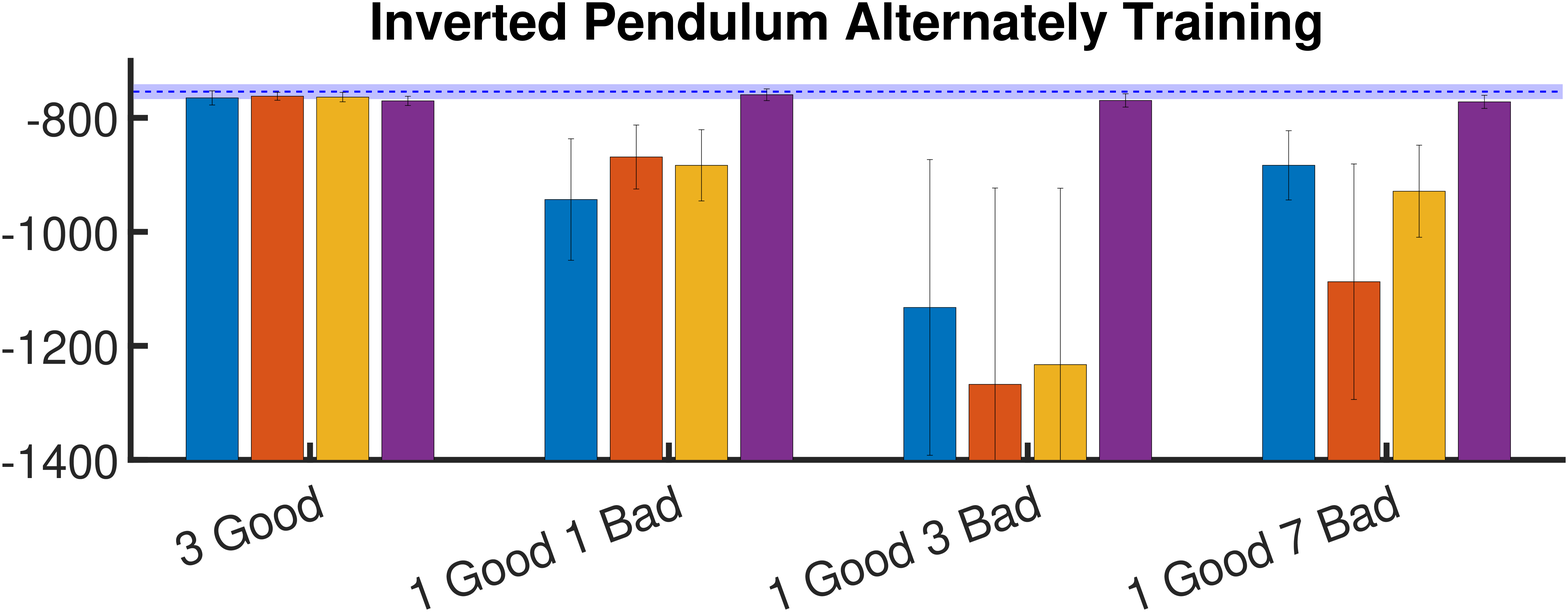}\hfill
            \includegraphics[width=.42\textwidth]{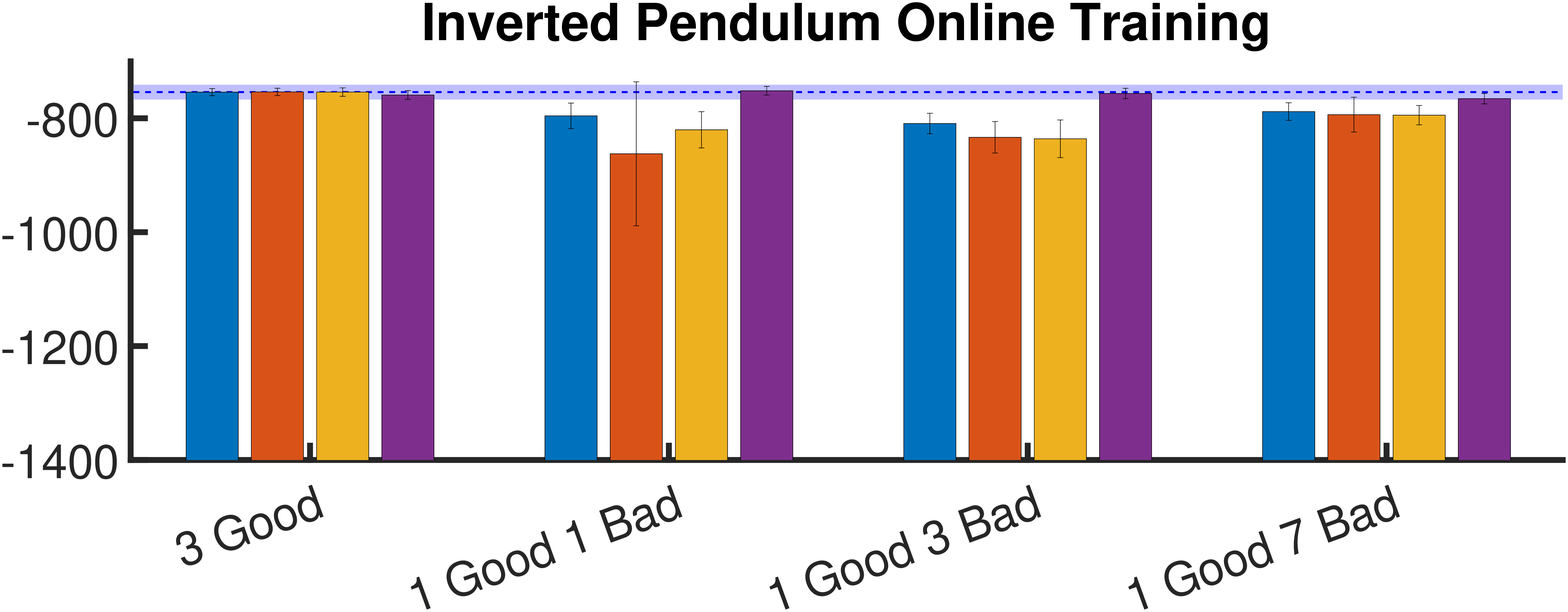}
            \\[\smallskipamount]
            \includegraphics[width=.42\textwidth]{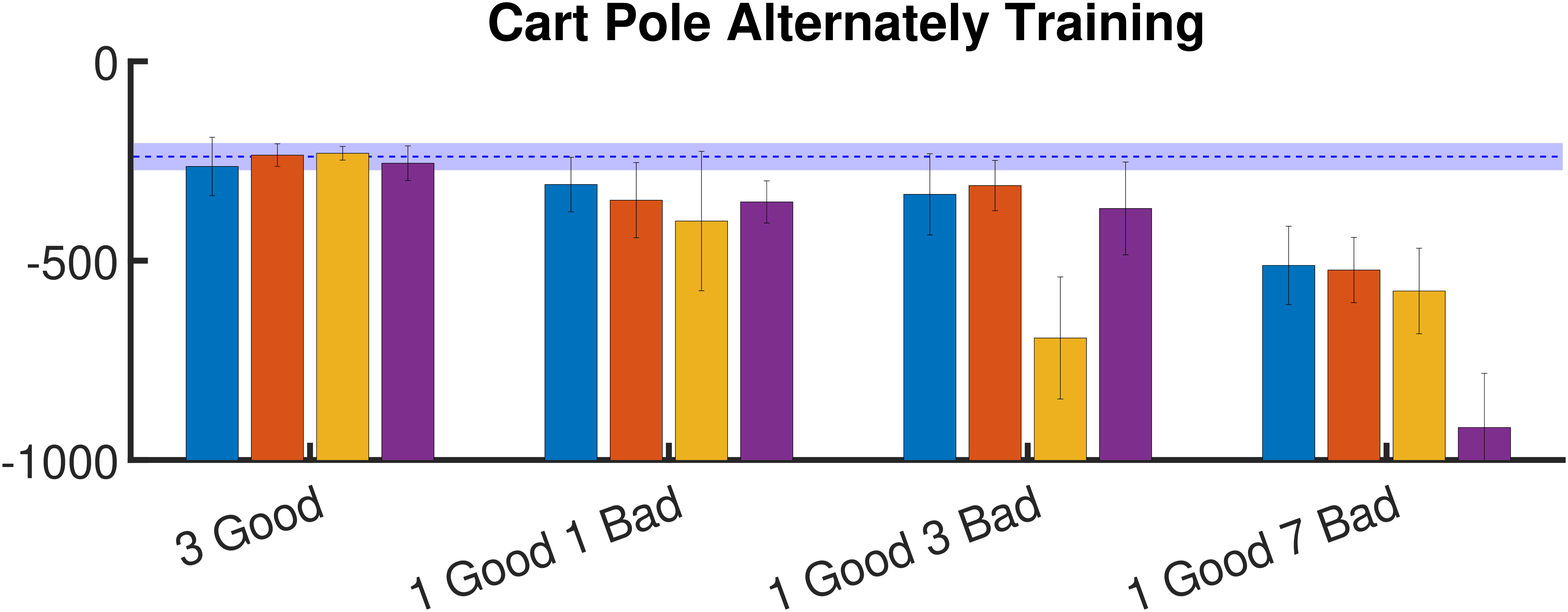}\hfill
            \includegraphics[width=.42\textwidth]{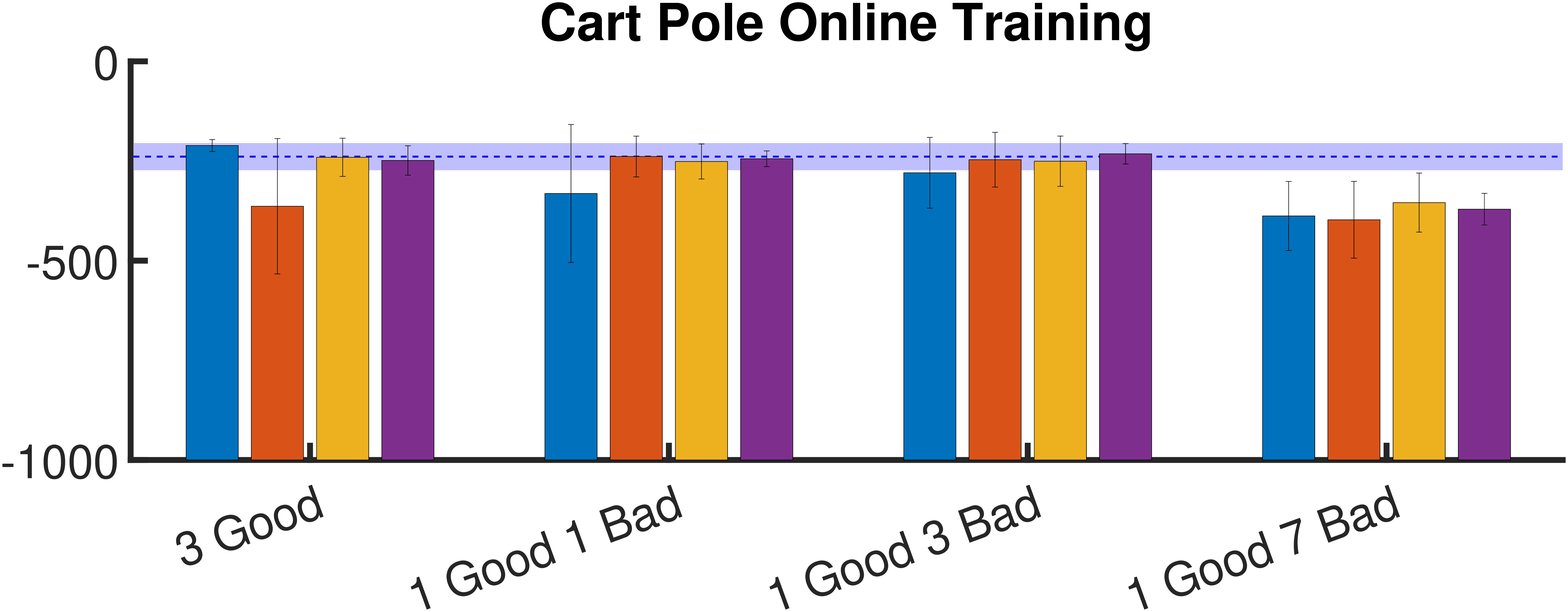}
            \\[\smallskipamount]
            \includegraphics[width=.42\textwidth]{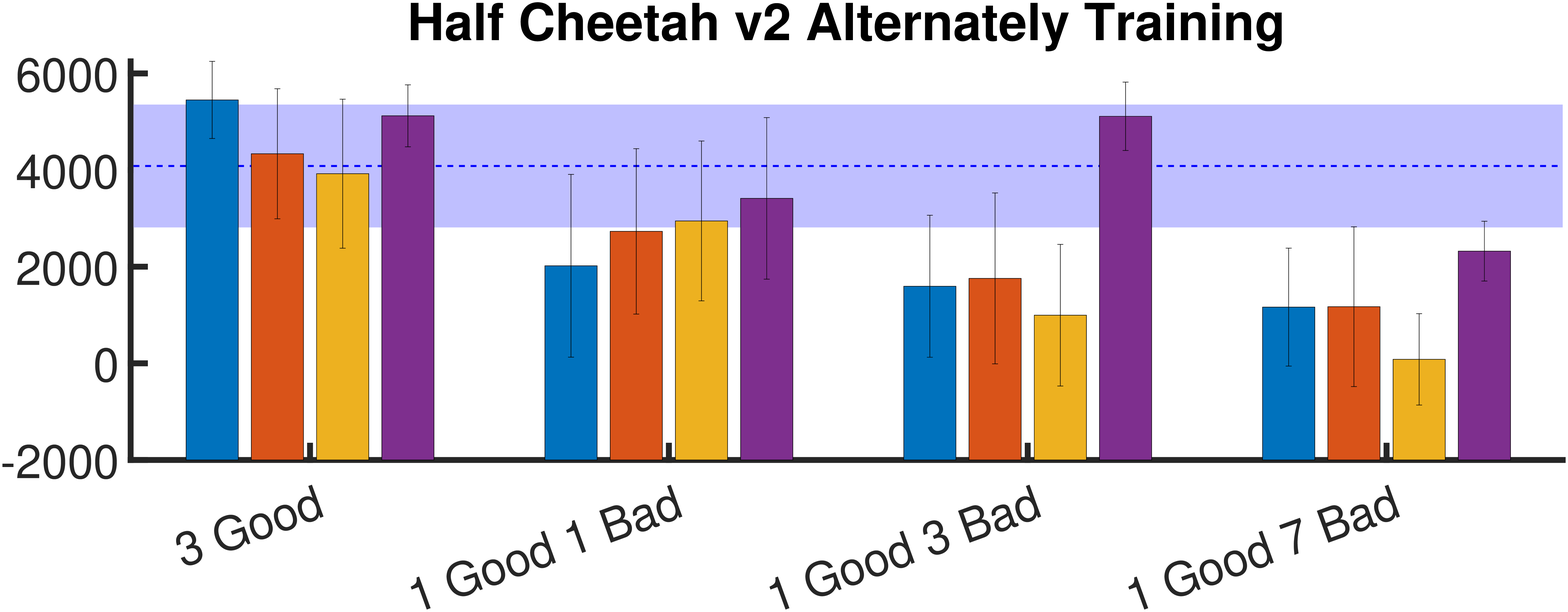}\hfill
            \includegraphics[width=.42\textwidth]{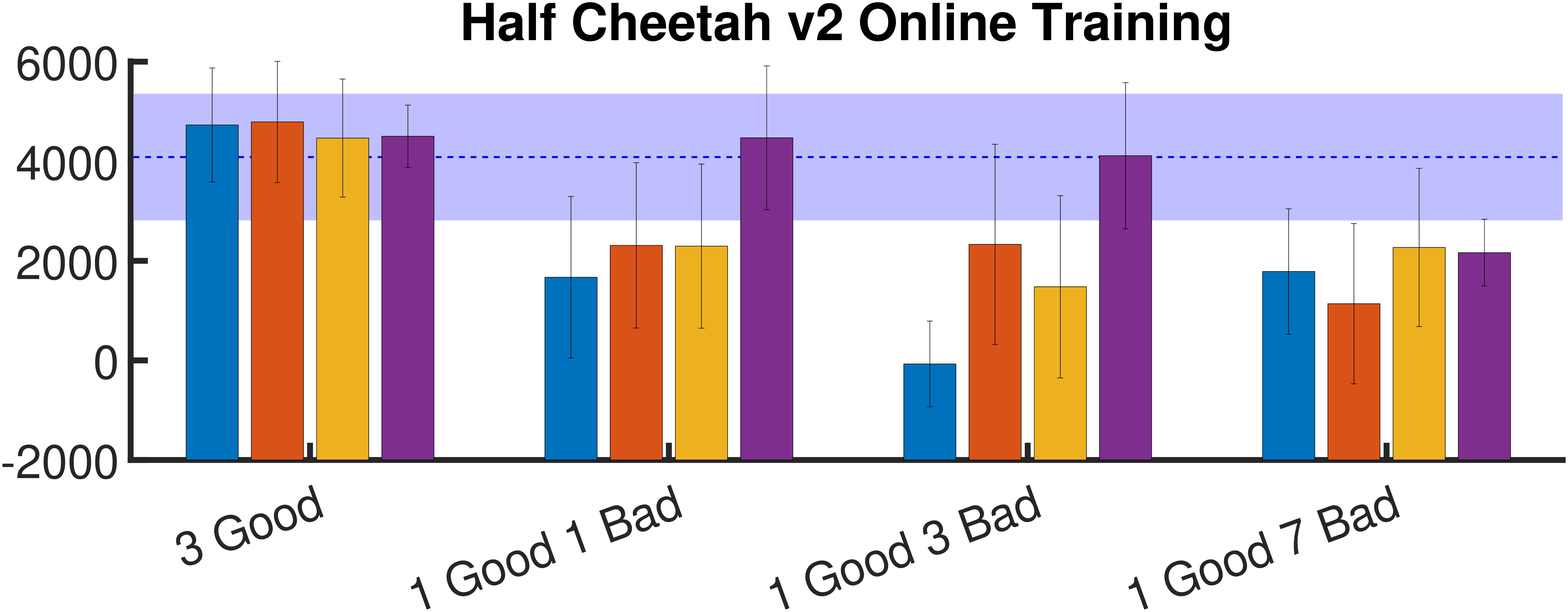}
            \\[\smallskipamount]
            \includegraphics[width=0.55\textwidth]{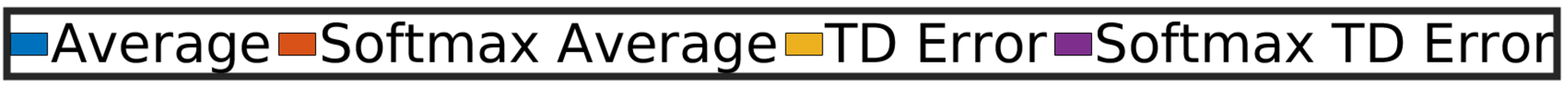}
            
            \caption{Performance and 95\% confidence interval of Online Weighted Q-Ensemble comparing groups (\textit{3 Best, 1 Good 1 Bad, 1 Good 3 Bad and 1 Good 7 Bad)} and Q-Aggregation methods (\textit{Average, Softmax Average, TD Error, Softmax TD Error}). Columns separate the Alternate and Online training and the rows present the environments: inverted pendulum, cart pole, and half cheetah v2. The horizontal line marks the mean of 30 (10 for half cheetah) single runs of the best parameterization, and the shadowed area represents its 95\% confidence interval.}
            \label{fig_performance_wce_bar_graphs}
        \end{figure*}

        Figure \ref{fig_performance_wce_bar_graphs} shows the final performance and its confidence interval. Each bar graph compares the performance on the 4 ensemble groups for the different aggregations, with separate graphs showing distinct environments and training modes. Also shown is the performance of the best single parameterization for each environment.
        
        We can observe that the online training mode almost always outperforms alternate training; furthermore, there is no significant variation between the aggregations in the \textit{3 Best} ensemble performance.
        Ensembles with a majority of bad parameterizations perform worse, which is especially evident in the more complex half cheetah v2  environment.
        
        In general, the \textit{Softmax TDError} aggregation performs better than, or within the confidence interval of, the other aggregations in the \textit{1 Good 3 Bad} group, and on par with the single and \textit{3 Best} ensemble even in the half cheetah v2  environment. As such, even when there are a majority of bad parameterizations in the ensemble, we can expect our proposed algorithm to perform similar to a single finetuned solution.

        Regarding the single ablations, there is no obvious trend as for which has the better performance.
        In fact, sometimes the intermediate strategies perform worse than simple Q-averaging. Clearly, both are required for optimum performance.
        
        Table \ref{tbl_average_relative_regret} presents the average relative regret of all strategies.
        Regardless of the training mode, the final model (\textit{Softmax TDError}) has a better evaluation, standing out in relation to the other strategies. 
        The latter rank differently depending on the training mode, making an impartial comparison between them impossible, thus it is not clear which aspect is more important for our model's final performance.
        
        
        The swimmer v2  validation uses 30 different ensembles, each with 8 randomly generated parameterizations.
        The performance of the full model (\textit{Softmax TDError} aggregation) was compared with simple Q-averaging, using online training.
        The mean and confidence intervals are $85\pm13$ for averaging, and $110\pm18$ for our model, showing a significant improvement.
        
        \begin{table}
    		\centering
    		\caption{Average relative regret using alternate and online training.}
    		\footnotesize
    		\label{tbl_average_relative_regret}
    		\begin{tabular}{|c|c|c|c|c|}
    			\hline
    			\multirow{2}{*}{\textbf{\textit{\vtop{\hbox{\strut Training }\hbox{\strut \ \ Mode}}}}} & \multirow{2}{*}{\textbf{\textit{Average}}} & \multicolumn{1}{c|}{\textbf{\textit{\vtop{\hbox{\strut Softmax}\hbox{\strut Average}}}}}  &
    			\multirow{2}{*}{\textbf{\textit{TDError}}} & \multicolumn{1}{c|}{\textbf{\textit{\vtop{\hbox{\strut Softmax}\hbox{\strut TDError}}}}} \\ \hline
    			 Alternate & 4.2280 & 4.7868 & 5.5208 & 2.5675 \\ \hline
    			 Online & 5.5059 & 6.0599 & 4.9232 & 2.3890 \\ \hline
    		\end{tabular}
    	\end{table}
        
    			             

	 \subsection{Learning curves}
    
\begin{figure*}
            \centering
            \includegraphics[width=.42\textwidth]{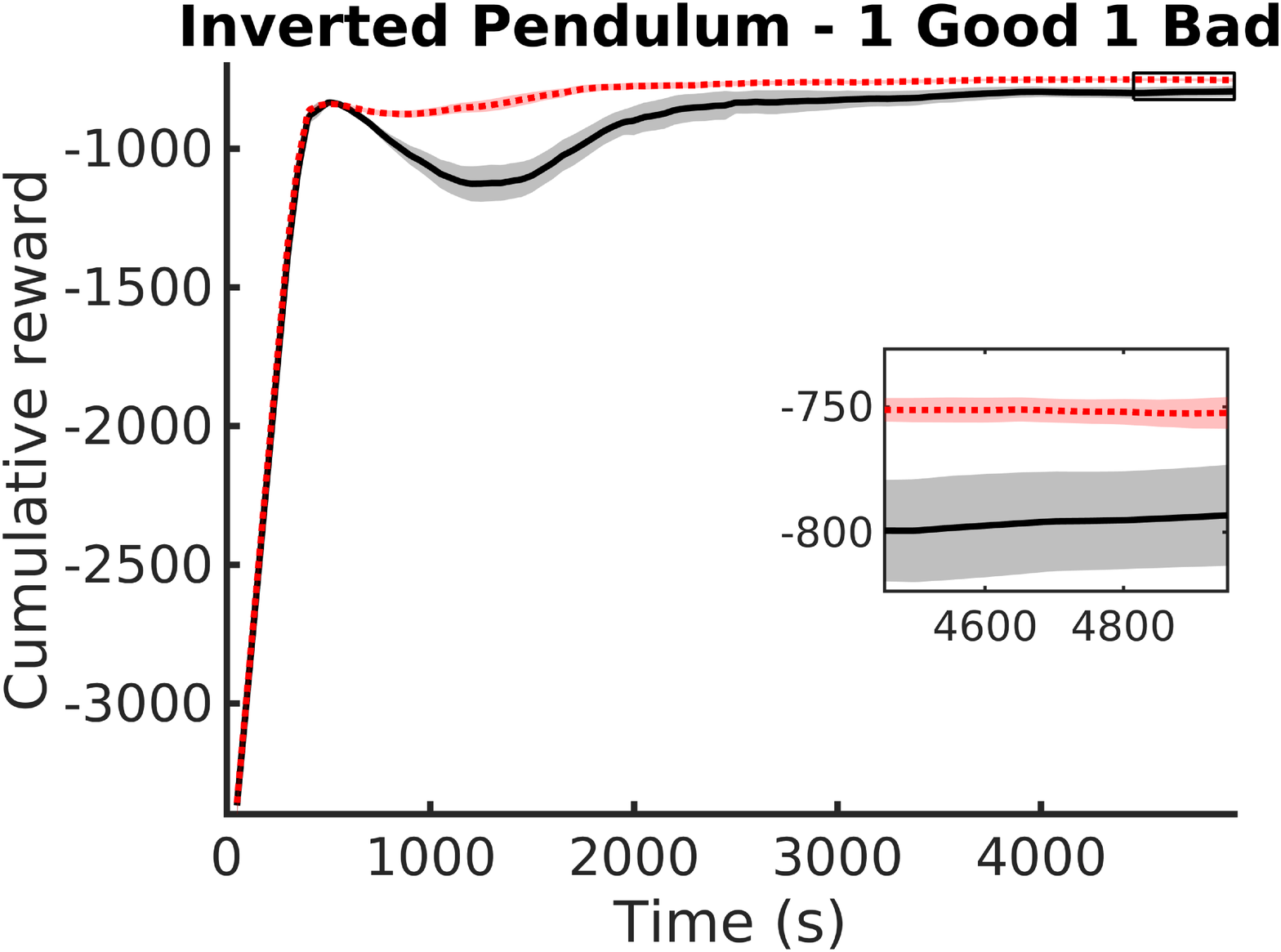}\hfill
            \includegraphics[width=.42\textwidth]{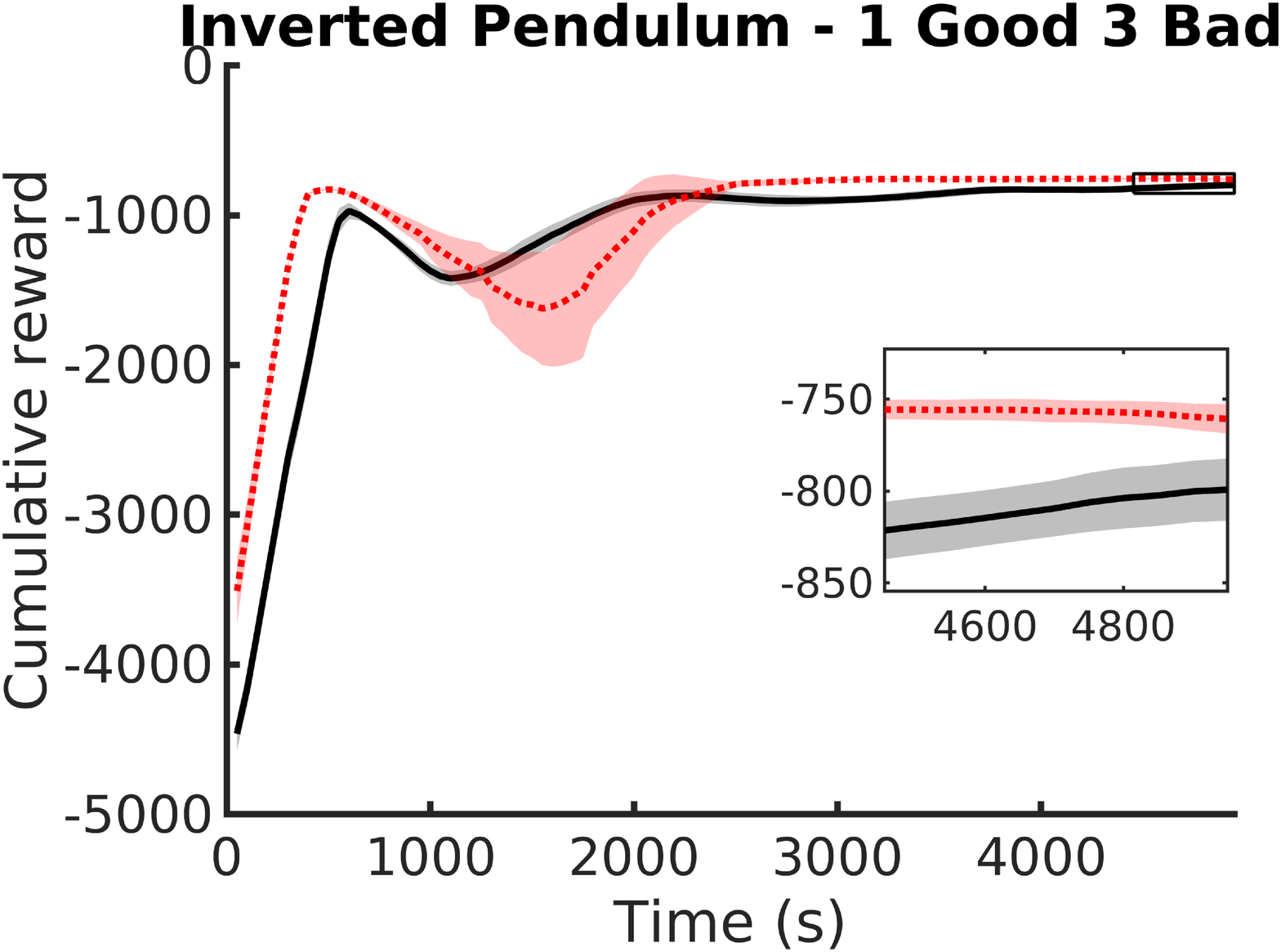}
            \\[\smallskipamount]
            \includegraphics[width=.42\textwidth]{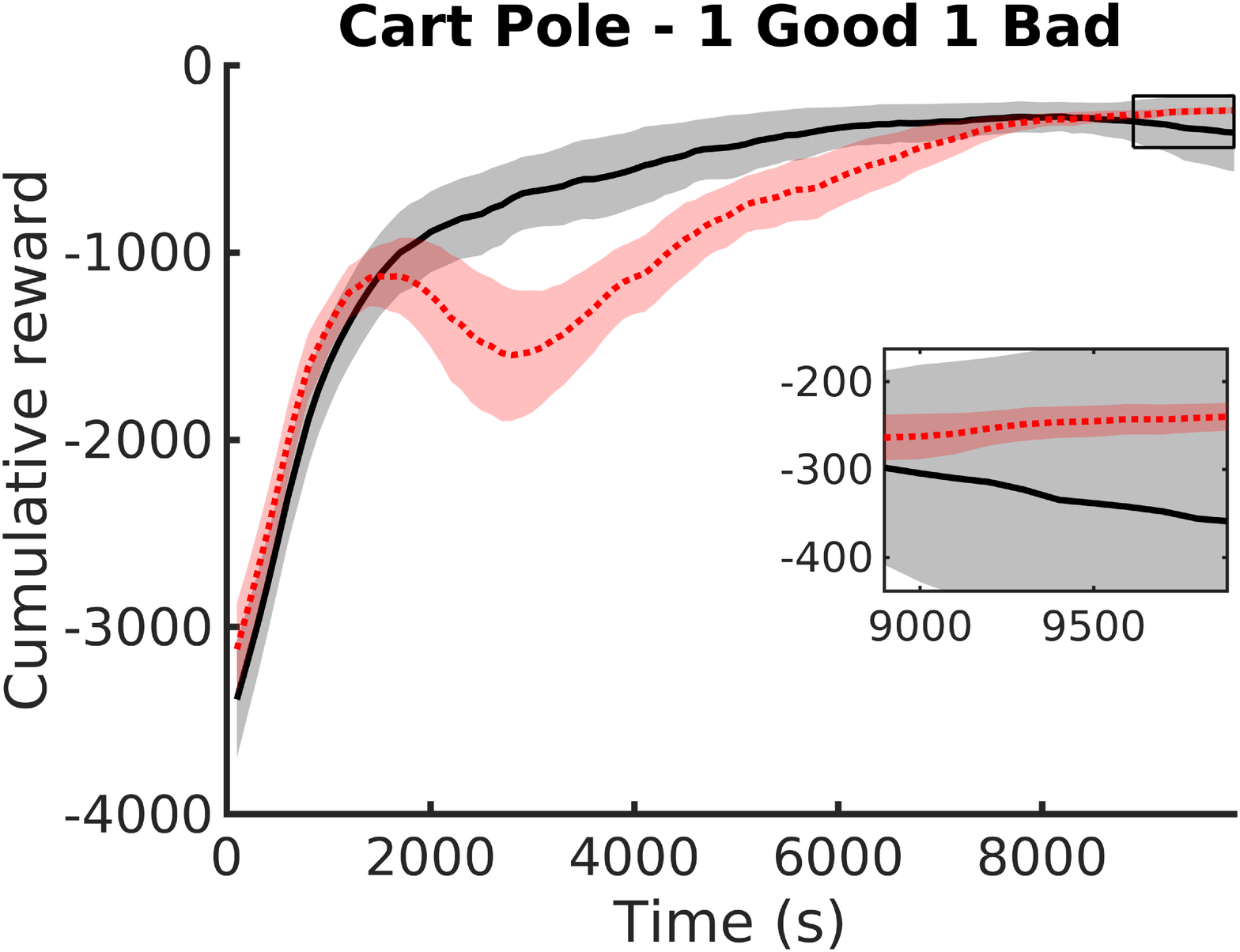}\hfill
            \includegraphics[width=.42\textwidth]{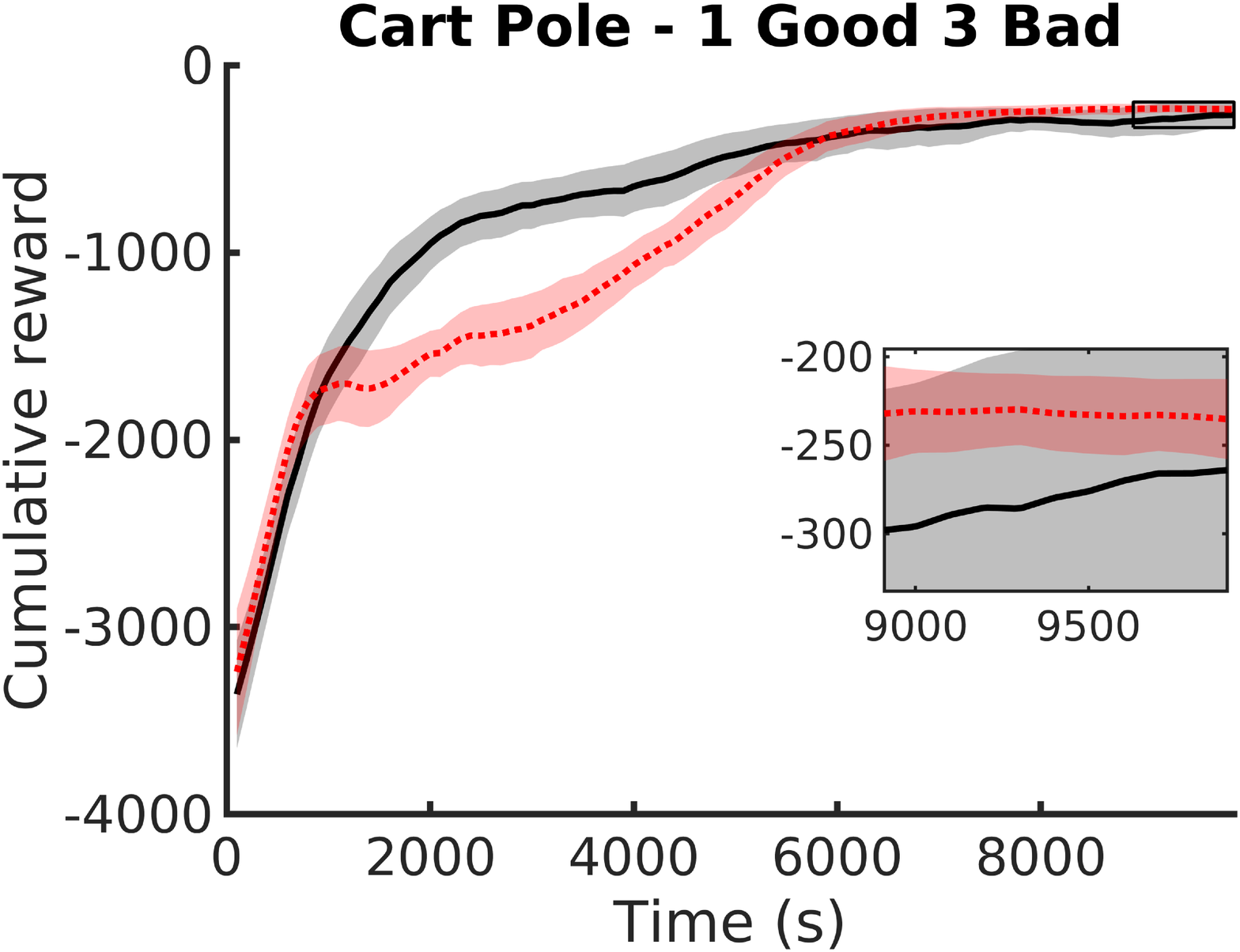}
            \\[\smallskipamount]
            \includegraphics[width=.42\textwidth]{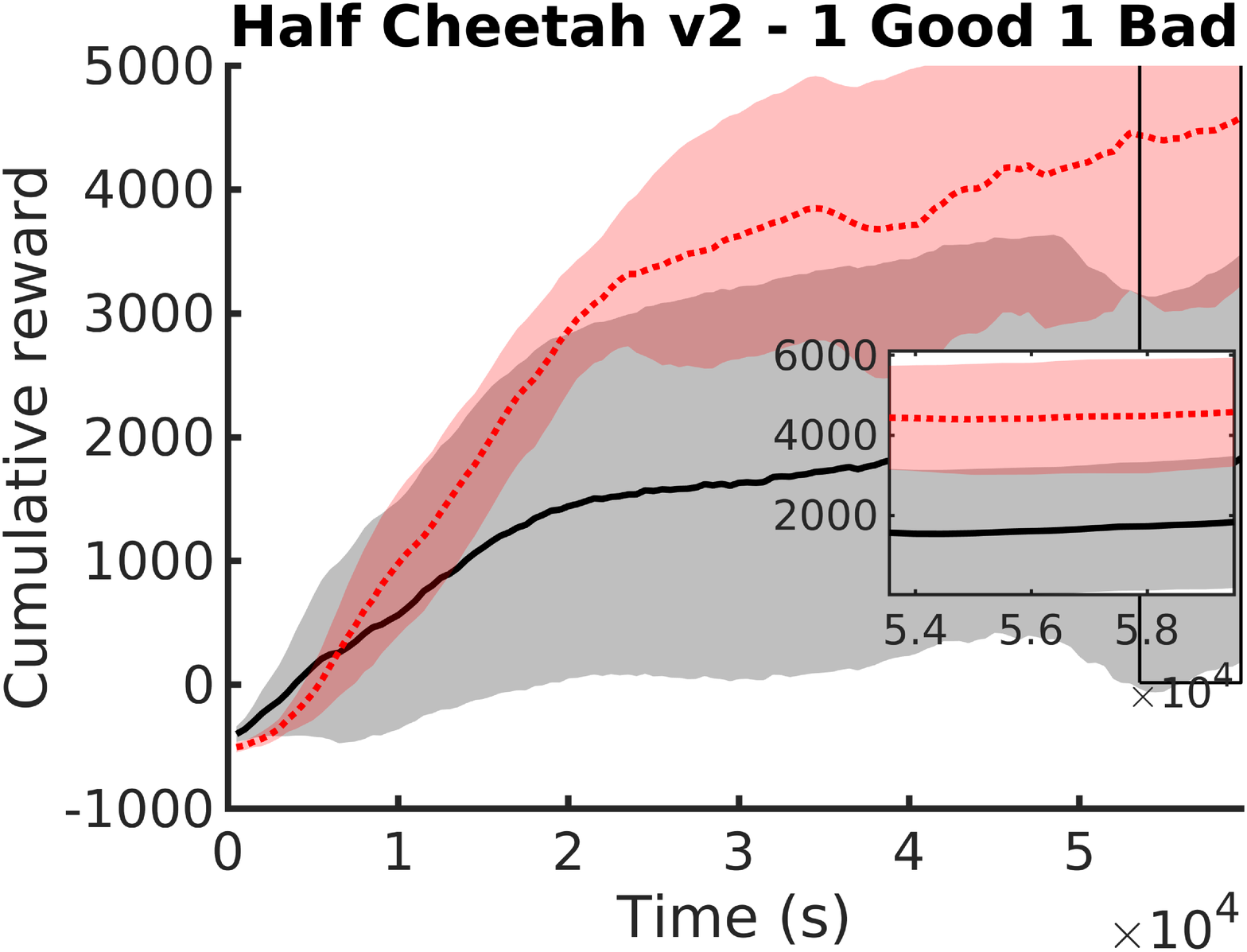}\hfill
            \includegraphics[width=.42\textwidth]{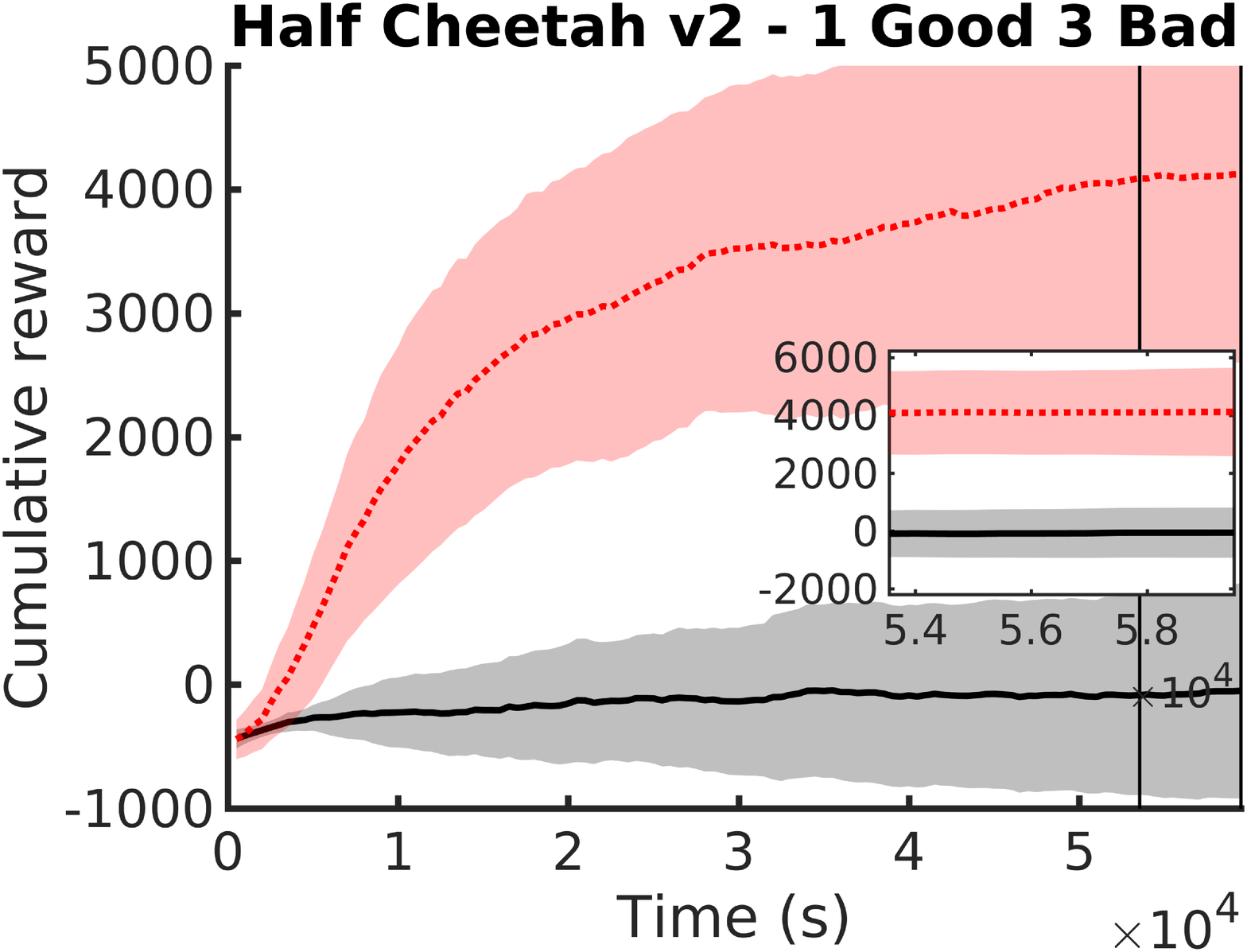}
            \\[\smallskipamount]
            \includegraphics[width=0.27\textwidth]{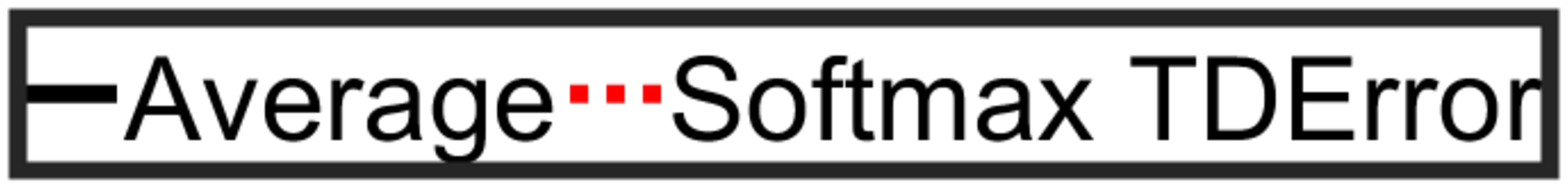}
            
            \caption{Learning Curve of Online Weighted Q-Ensemble with \textit{1 Good 1 Bad, and 1 Good and 3 Bad} and Q-Aggregation (\textit{Average, Softmax TD Error}).  All cases are online training, lines present the environments: inverted pendulum, cart pole and half cheetah v2. The graphic also shows the 95\% confidence interval.}
            \label{fig_performance_wce_learning_curve}
        \end{figure*}

        Figure \ref{fig_performance_wce_learning_curve} shows the online training mode learning curves of the \textit{1 Good 1 Bad} and \textit{1 Good 3 Bad} ensembles for the \textit{Average} and \textit{Softmax TDError} aggregations. In the inverted pendulum, \textit{Softmax TDError} learns a bit faster than \textit{Average}, while in the cart-pole this is behavior is reversed. In both environments learning is stable with good end performance, with \textit{Softmax TDError} having higher mean and lower final variance. 

        In half cheetah v2 , the performance difference is huge. Both curves show high variance, but in both ensemble groups \textit{Softmax TDError} performs better. Specifically, \textit{Average} does not manage to learn in the \textit{1 Good 3 Bad} ensemble, while \textit{Softmax TDError} maintains the same performance as in the \textit{1 Good 1 Bad} group.
        
	 \subsection{Action preference and Q-weights}
        
        \begin{figure*}
            \centering
            \includegraphics[width=.42\textwidth]{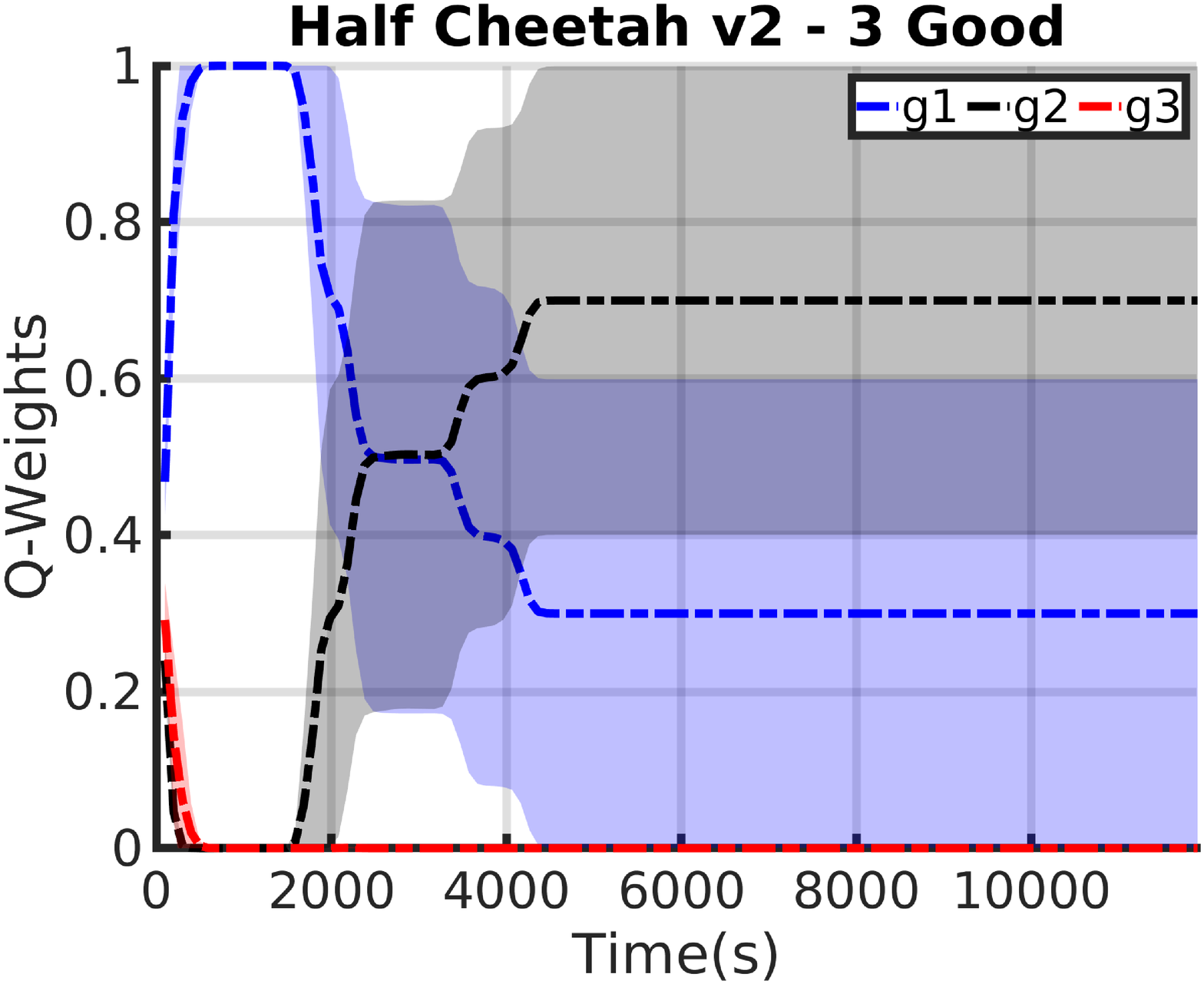}
            \includegraphics[width=.42\textwidth]{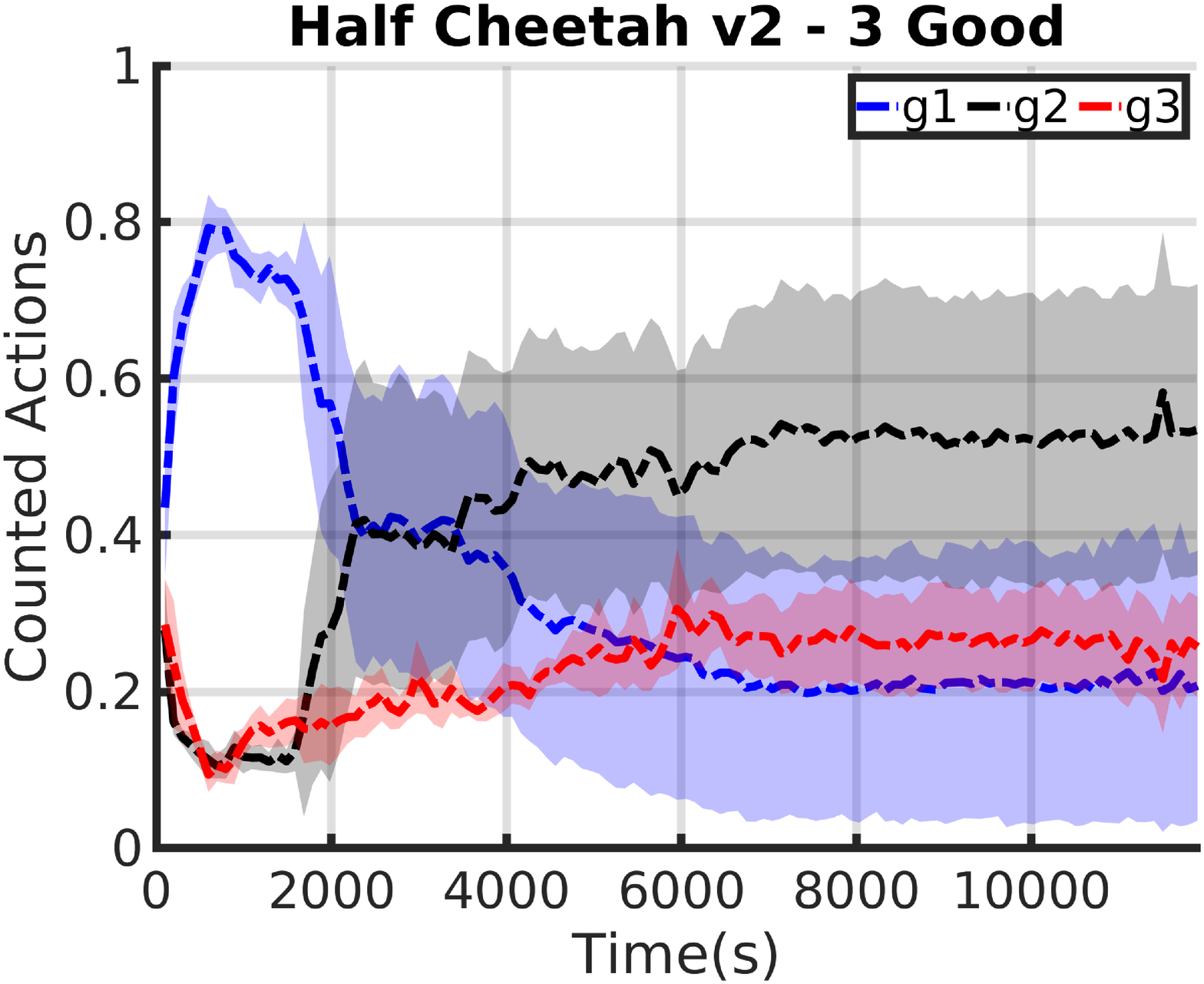}
            \\[\smallskipamount]
            \includegraphics[width=.42\textwidth]{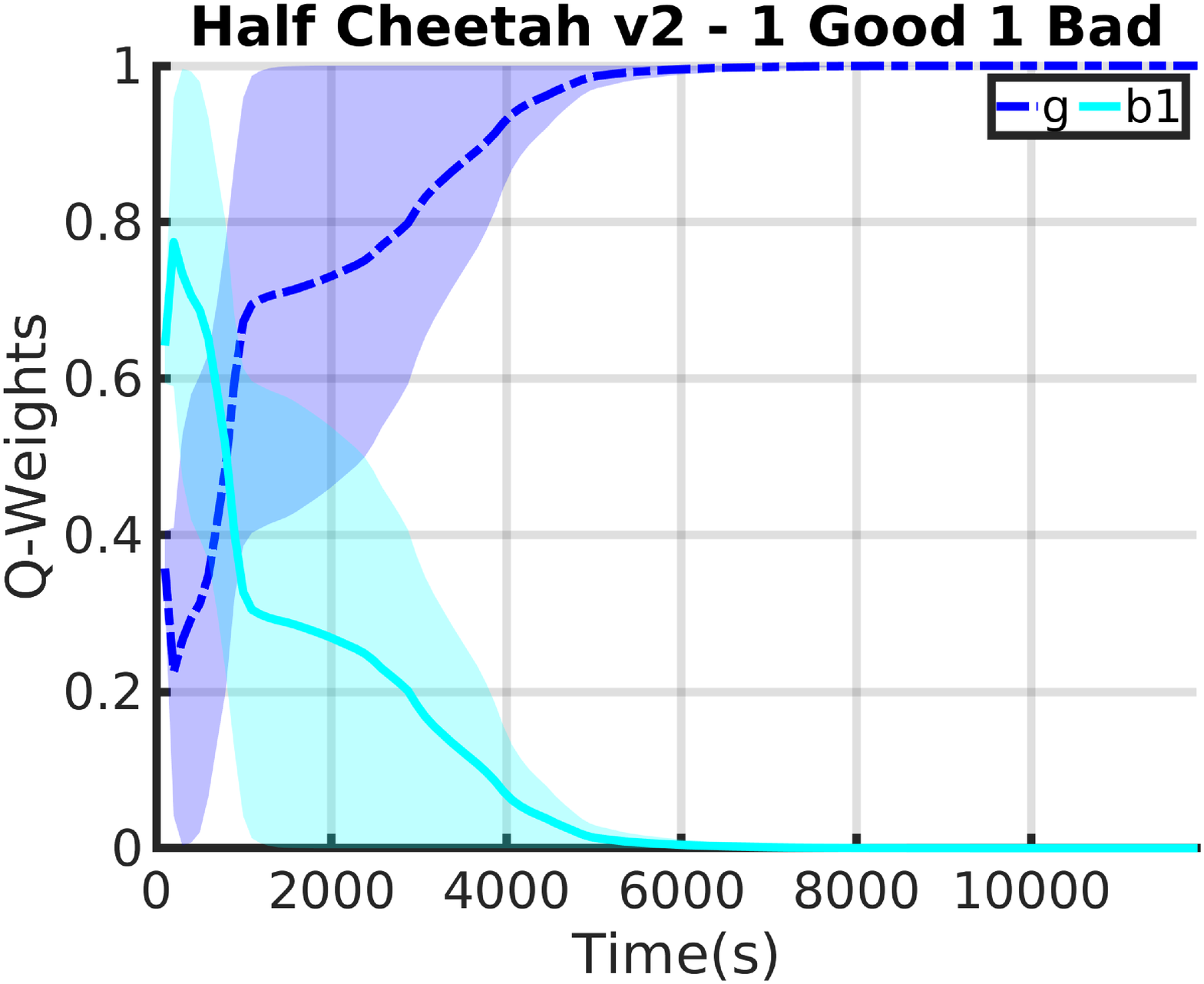}
            \includegraphics[width=.42\textwidth]{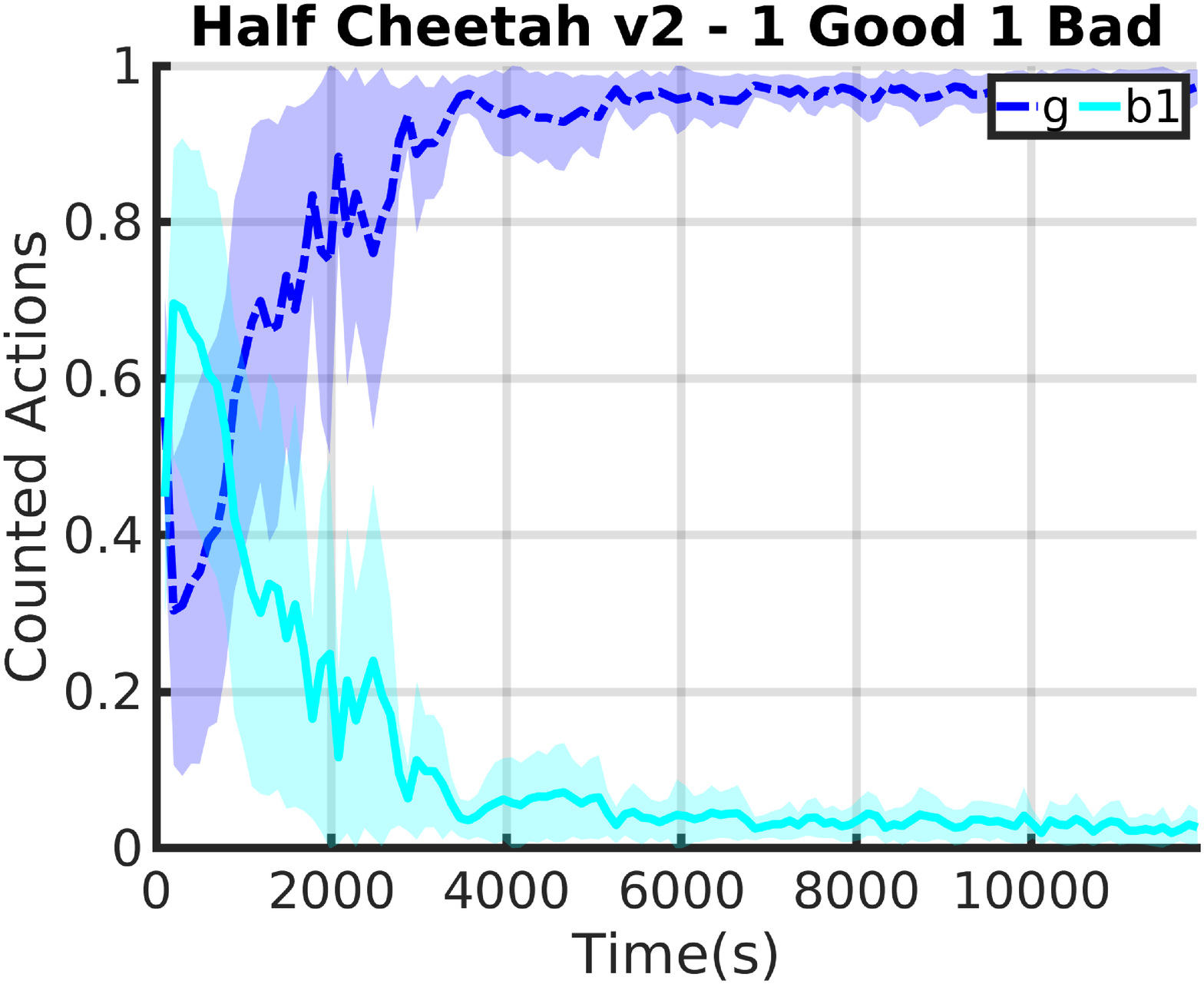}
            \\[\smallskipamount]
            \includegraphics[width=.42\textwidth]{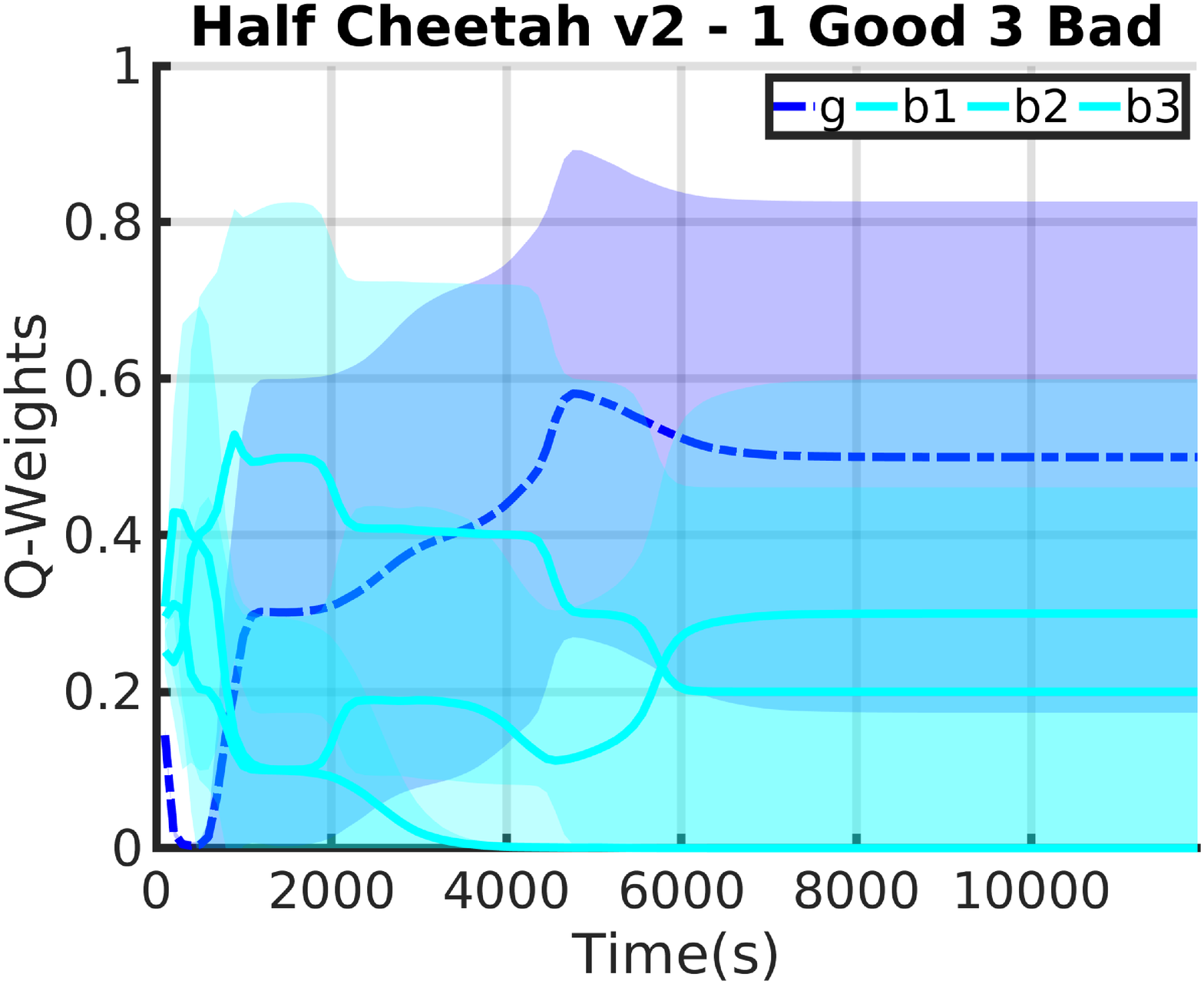}
            \includegraphics[width=.42\textwidth]{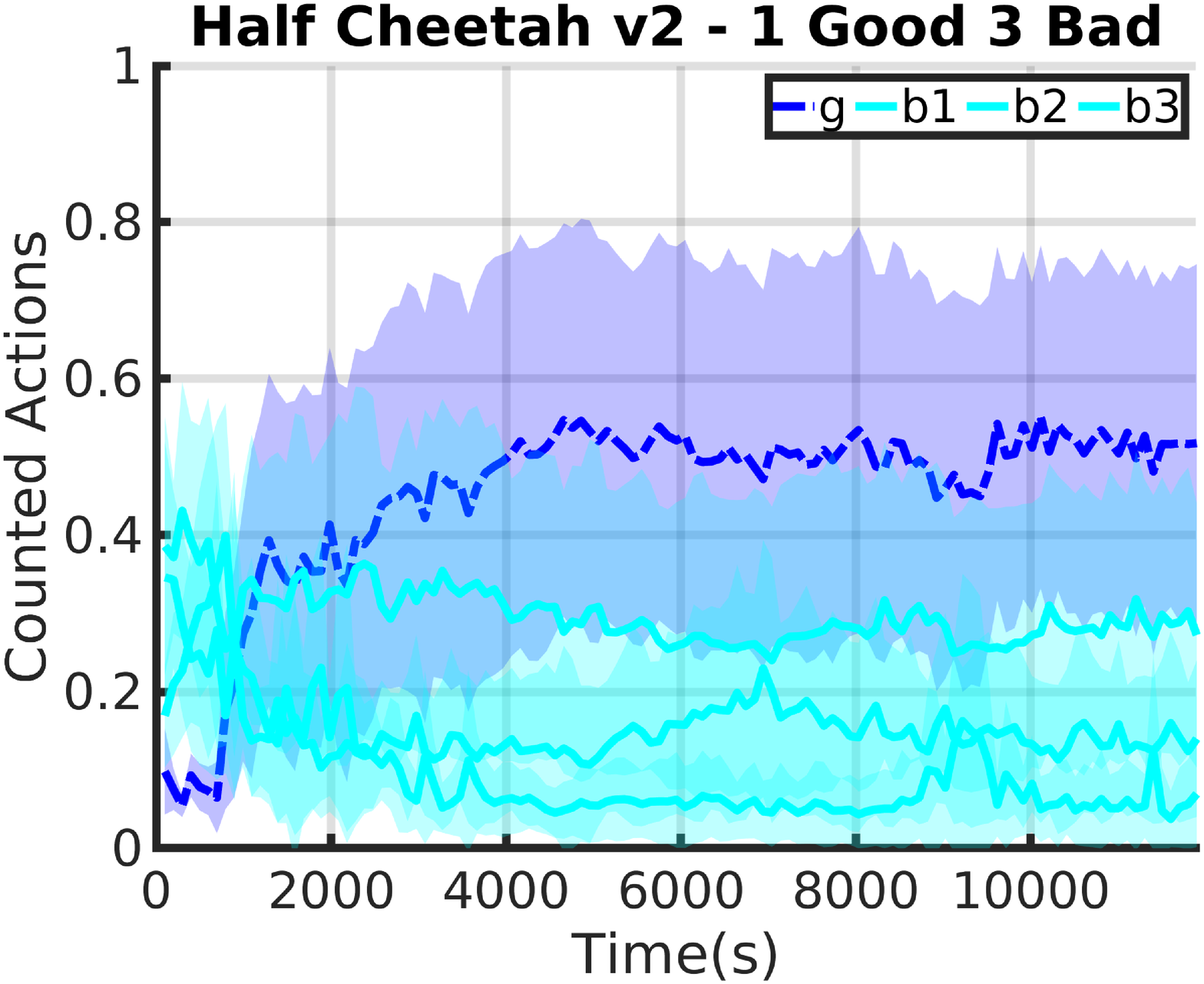}
            \\[\smallskipamount]
            \includegraphics[width=.42\textwidth]{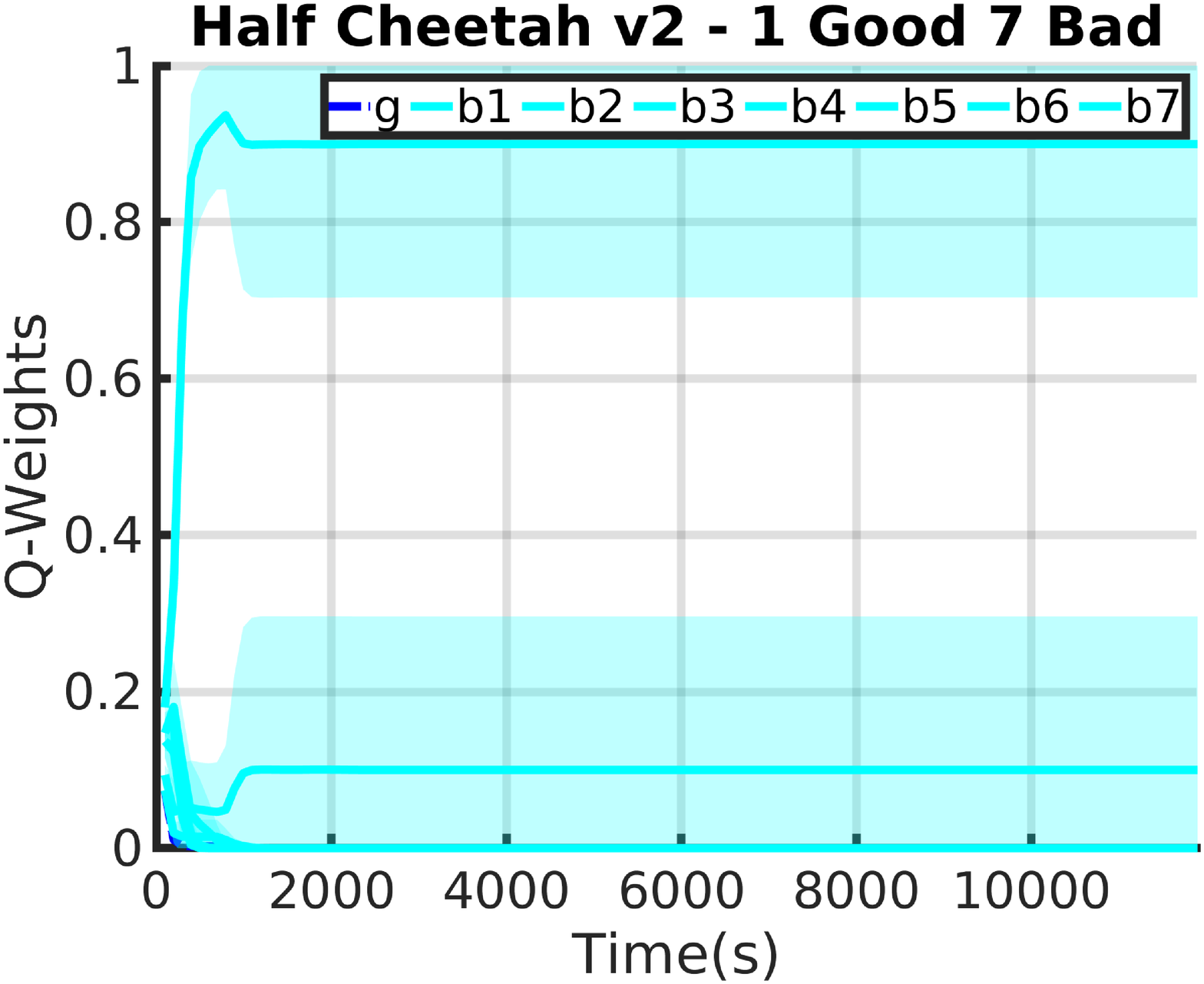}
            \includegraphics[width=.42\textwidth]{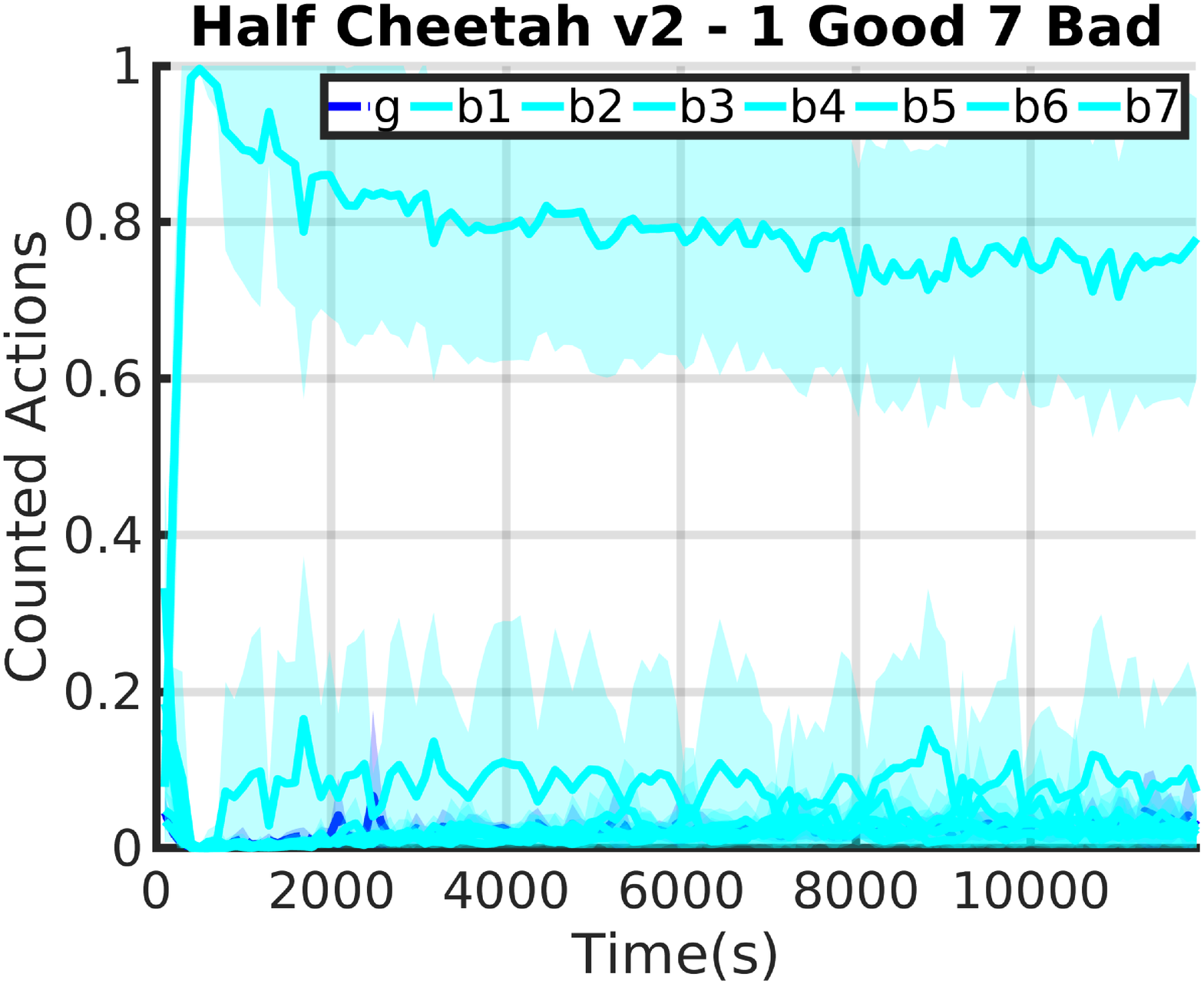}
            \\[\smallskipamount]
            \caption{Counter Action and Q-Weights of {\OWQEext} with \textit{3 Best, 1 Good 1 Bad, 1 Good and 3 Bad, and 1 Good and 7 Bad} and Q-Aggregation \textit{Softmax TD Error}). All cases are online training in half cheetah v2  environment. The graphic also shows the 95\% confidence interval.}
            \label{fig_wce_counter_action_q_weights}
        \end{figure*}
	 
        To better understand our model's behavior, an investigation of how the actions are chosen was made.
        Figure \ref{fig_wce_counter_action_q_weights} presents the behavior of the half cheetah v2  MuJoCo environment, where the left column shows the weights assigned to each critic ($\mathbf{W}$), and the right column shows how often the different actors' actions were chosen in the episode.
        All bad parameterizations have the same color, while the good ones are highlighted.
        
        At the beginning of the learning process, $\mathbf{W}$ is uniformly distributed, and at the end the weights tend to choose the critic with the lowest calculated TD Error.
        The same process happens with the counted actions; the beginning of the learning process has an equal distribution of the chosen actions, while at the end the choice of actions is influenced by the quality of the policy and the critics' acquired Q-Weights.
        
        The first row shows the \textit{3 Good} group, where each parameterization has a different color.
        Intuitively, the expectation is that both weights and action choices should stay equally distributed, since all parameterizations have roughly the same individual performance.
        Initially, this is not the case, as some choices may learn faster than others, but, at least for the actions, the end result is as expected.
        Note that one critic has a very low weight, but its actor's action is chosen normally (red line).
        This shows that having a higher TD-error critic does not always imply a worse actor.

        The second row presents the \textit{1 Good 1 Bad} group, which is composed of two opposing parameterizations, with an expectation that the good parameterization will be chosen from the beginning.
        Indeed, there is a very clear distinction from the Q-Weights, that is reflected in the choice of actions with little noise.
        The third row of Figure \ref{fig_wce_counter_action_q_weights} presents the \textit{1 Good and 3 Bad} group, which behaves similarly to the \textit{1 Good 1 Bad}, but with more challenges, as there are more bad parameterizations to compete with.
        
        Finally, the \textit{1 Good and 7 Bad} group shown in the last row struggles to find the good agent.
        The Q-Weights do not manage to converge to the best individual critic, nor is its action chosen more often than the others. However, the results in Figure~\ref{fig_performance_wce_bar_graphs} show that the ensemble still reaches an adequate (although not optimal) performance.  
        
        
        Overall, the best individual agent has both lower weight and its actions are generally chosen less in larger ensembles. Even so, there is an improvement in the final performance when \textit{Softmax TDError} is used.

\section{Conclusion}
\label{sec_conclusion}

	This article proposed the Online Weighted Q-Ensemble to decrease the hyperparameter tuning effort for deep reinforcement learning in continuous action spaces.
	Based on previous work which uses an average of Q-ensembles in an actor-critic setting, we introduced a weighing approach that adjusts the critics' weights by minimizing the temporal difference error of the ensemble as a whole.
	Additionally, instead of combining the Q-values directly, they were applied through a softmax layer, in order to focus on relative preferences rather than absolute values.
	
	In both simple and complex robotic simulation environments, our model showed better results than the standard Q-value averaging, and managed to maintain performance comparable to the best individual run even if the ensemble included up to 3-7 bad parameterizations.
	Validation using ensembles with 8 randomized parameterizations also showed superior performance compared to q-value averaging.
	
	Our tests used a single environment, as they were aimed at the system's applicability in real-world robotic applications. In future work, it would be interesting to extend the simulations to more environments, and validate its use in real robots.
	Other interesting points to be further expanded in possible subsequent works are the acceleration of learning and the extension of tests with further algorithms, such as TD3 and SAC.
    Furthermore, while performance with mostly bad parameterizations was adequate, algorithmic improvements could be made to further suppress their influence, increasing robustness and decreasing the need to select good hyperparameter ranges.
    Finally, extensions to include the discount rate and reward scale in the tunable hyperparameters could be considered.

\section*{Acknowledgements}

This study was financed in part by the Coordenação de Aperfeiçoamento de Pessoal de Nível Superior - Brasil (CAPES) - Finance code 001 and the National Council for Scientific and Technological Development -- CNPq under project number 314121/2021-8.



\bibliographystyle{plain}
\bibliography{article}


\end{document}